\documentclass[preprint,12pt,authoryear]{elsarticle}

\journal{Ocean Engineering}

\usepackage{amsmath,amsfonts,amssymb,mathtools}
\usepackage{algorithmic}
\usepackage{algorithm}

\usepackage{array}
\usepackage[font=normalsize,labelfont=bf]{caption}
\usepackage{subfig}
\usepackage{textcomp}
\usepackage{url}
\usepackage{verbatim}
\usepackage{graphicx}
\usepackage[T1]{fontenc}
\usepackage{float}
\usepackage{placeins}
\usepackage{diagbox}
\usepackage{tikz}
\usepackage{pgfplots}
\usepackage{booktabs}
\usepackage{multirow}
\usepackage{makecell}
\pgfplotsset{compat=1.18}
\usetikzlibrary{arrows.meta,positioning,fit,calc,backgrounds,shapes.geometric}
\biboptions{authoryear,round}
\begin{document}

\begin{frontmatter}

\title{Knowledge-Distilled End-to-End Reinforcement Learning for Smooth 6-DOF Thrust Control and Rapid Adaptation to Ocean Currents in Remotely Operated Vehicles}

\author[a]{Tiankuang Wen}

\author[a]{Huiping Li\corref{cor1}}
\ead{lihuiping@nwpu.edu.cn}

\author[b]{Gang Liu}

\author[b]{Yong Jiang\corref{cor1}}
\ead{jiangyongbobby@sina.cn}

\cortext[cor1]{Corresponding author.}

\affiliation[a]{
	organization={Northwestern Polytechnical University},
	city={Xi'an 710129},
	country={China}
}

\affiliation[b]{
	organization={Wuhan Second Ship Design and Research Institute},
	city={Wuhan 430205},
	country={China}
}

\begin{abstract}
With the continuous improvement of computational capabilities, end-to-end reinforcement learning has been rapidly developed for remotely operated vehicles control. Nevertheless, existing end-to-end reinforcement-learning-based methods still face challenges in achieving optimal control under ocean-current disturbances. In particular, there remains a lack of a unified control framework that can simultaneously achieve low steady-state tracking error, rapid transient response, energy-efficient operation, and smooth control-force outputs under disturbances. To address the issue, this paper proposes the thrust smoothness rapid current adaptation proximal policy optimization (TSRCA-PPO) method which learns a near-optimal strategy by a two-stage distillation learning framework. The core innovations of this work lie in the reward-function design and the privileged multi-encoder architecture.
 Ablation studies validate the effectiveness of each module. 
Simulation results demonstrate that the proposed TSRCA-PPO method consistently
outperforms the conventional cascaded P-PID controller across all evaluation
metrics. Specifically, TSRCA-PPO reduces the steady-state position error, steady-state attitude error,
settling time, energy index, and thrust-smoothness index
to \(42.7\%\), \(76.5\%\), \(10.6\%\), \(93.5\%\), and \(15.9\%\) of the
corresponding P-PID values, respectively.
\end{abstract}


\end{frontmatter}

\section{Introduction}
Remotely operated vehicles (ROVs) are increasingly used in marine exploration, inspection, and intervention tasks, thanks to their high maneuverability and precise control capabilities~\citep{gao2024dynamic,deng2025thrust}. To accomplish such missions, they are often required to reach prescribed target locations and maintain station for extended periods, particularly under persistent currents. Furthermore, since propulsion effort directly contributes to energy consumption and mechanical wear, reducing control effort and suppressing control chattering can lower operational costs and extend mission endurance~\citep{walters2018online}.

A common control architecture for ROVs systems is hierarchical: a high-level motion controller first generates desired virtual control commands, such as generalized forces and moments, and a low-level control allocation module then maps these commands into individual thruster inputs. This modular architecture allows the high-level motion controller to be designed without detailed knowledge of the effector and actuator system. However, the interactions between motion control and control allocation should be studied or explicitly accounted for when actuator dynamics are significant~\citep{johansen2013control}, such as for T200 thrusters, commonly used in small UUVs and ROVs, which exhibit non-negligible input dead zones, saturation, and nonlinear thrust generation characteristics~\citep{lam2023t200}. To address the limitations, integrated control-allocation approaches have been proposed, in which motion control and control allocation are designed jointly rather than treated as two independent modules~\citep{barlund2020nonlinear,martinsen2022reinforcement,shen2025integrated,overeng2021dprl,tuncay2026fastpolicy,cai2025swim,sufan2025swim4real,chu2025marinegym}.

 Previous studies on integrated motion-control and thrust-allocation methods have mainly focused on model predictive control (MPC)-based approaches, as MPC provides a natural framework for jointly handling motion-control objectives, actuator constraints, and thrust allocation~\citep{barlund2020nonlinear,martinsen2022reinforcement,shen2025integrated}. Although integrated MPC-based approaches have achieved promising performance, most existing methods are developed and validated using simplified vehicle dynamics. Extending these approaches to full six-degree-of-freedom underwater vehicle dynamics would substantially increase the size and complexity of the online optimization problem, posing significant challenges for real-time deployment. Moreover, their reliance on accurate system models remains a key limitation, since underwater hydrodynamics are highly nonlinear, uncertain, and difficult to identify precisely in real operating conditions~\citep{wei2023mpc}.

Recent advances in deep reinforcement learning provide a promising alternative for addressing these challenges by shifting the main computational burden from online constrained optimization to offline trial-and-error policy learning in simulation. As a result, real-time 6-DOF integrated control-allocation methods can be achieved through fast policy inference. Furthermore, reinforcement-learning-based methods reduce the reliance on accurate analytical models, but they transfer this burden to the development of realistic simulation environments and effective sim-to-real adaptation strategies~\citep{tang2024deep}. Sufficient exposure to diverse hydrodynamic uncertainties, actuator dynamics, sensor noise, and environmental disturbances can therefore improve the vehicle's adaptability to model mismatches and unseen operating conditions.
 
Building on these developments, an increasing number of studies have investigated end-to-end actuator-level reinforcement learning for marine vehicle control. In this paradigm, the policy directly maps vehicle states or observations to actuator commands, thereby integrating motion control and thrust allocation within a single learned controller. \citet{carlucho2018adaptive} proposed a DDPG-based end-to-end low-level control method for AUV velocity tracking. By jointly penalizing tracking error, thruster usage, and variations in control actions, the method learned a relatively accurate control policy with low control effort and smooth thruster commands. Its feasibility was subsequently demonstrated through simulations and real-world pool experiments using the Nessie VII AUV. \citet{overeng2021dprl} proposed a Proximal Policy Optimization (PPO)-based neural-network policy that encapsulates motion control and control allocation for dynamic positioning of marine surface vessels. An integral state augmentation was introduced to reduce steady-state error in simulation; However, oscillations and overshoot remained near the steady state under a constant ocean-current disturbance. \citet{cai2025swim} proposed a fast 6-DOF reinforcement learning framework that maps command-conditioned observations directly to low-level thruster commands. Using massively parallel simulation and domain randomization, their policy achieved zero-shot transfer to a real AUV and showed robustness to moderate variations in physical parameters. Nevertheless, real-world deployment in station-keeping tests still exhibited steady-state errors. \citet{sufan2025swim4real} further emphasized energy-efficient 6-DOF underwater control using the Truncated Quantile Critics (TQC) algorithm, achieving accuracy comparable to a well-tuned PID controller while reducing energy consumption in controlled calm-water tank experiments. Meanwhile, MarineGym provides a high-performance benchmark platform for underwater RL research, facilitating end to end reinforcement learning training~\citep{chu2025marinegym}. However, among the aforementioned studies, only \citep{chu2025marinegym} explicitly considered varying ocean-current disturbances. Furthermore, it neither conducted a comprehensive comparison of different algorithms nor proposed a control architecture specifically tailored to operation under ocean-current disturbances. Some non-end-to-end methods account for external ocean-current disturbances~\citep{luo2025rovadrc,huang2023general}. However, as discussed above, when actuator dynamics are significant, generalized force and moment generation, control allocation, and individual thruster command generation should be considered jointly; otherwise, discrepancies between the commanded and delivered thrust may degrade control performance.

Despite the considerable potential of end-to-end reinforcement-learning-based methods for underwater vehicle control, achieving optimal control under ocean-current disturbances remains challenging. First, the vehicle must simultaneously minimize multiple performance objectives,
including tracking errors, convergence time, energy consumption, and thrust
fluctuations in the presence of observation noise, ocean-current disturbances, and uncertainties in its internal parameters. Furthermore, reinforcement-learning-based policies are prone to generate oscillatory thruster-level commands, making it more difficult to obtain optimal control actions. ~\citep{song2025lipsnet,lee2024gradient,mysore2021regularizing}. 

To solve these issues, we propose the thrust smoothness rapid current adaptation PPO (TSRCA-PPO) algorithm. To ensure optimal overall performance, we carefully designed a reward function comprising linear-velocity guidance, angular-velocity guidance, energy regularization, smoothness regularization, terminal position, and terminal attitude rewards. In particular, the velocity-guidance terms encourage the vehicle to decelerate as it approaches the target, thereby reducing overshoot and oscillations. To enable the policy to respond rapidly to changes in ocean currents while generating smooth thrust commands that reduce actuator wear, we introduce a privileged multi-encoder architecture that exploits global information to optimize control performance. Subsequently, to decouple optimal policy learning from implicit state estimation, inspired by \citet{kumar2021rma,xiao2024paloco}, we employ policy distillation to transfer the optimal policy learned with privileged information to a deployable student policy. The main contributions of this study are summarized as follows:
\begin{itemize}
\item A two-stage knowledge-distillation-based reinforcement learning framework, named TSRCA-PPO, is proposed to learn a near-optimal policy for underwater vehicle position control.
\item A novel reward function comprising linear-velocity guidance, angular-velocity guidance, energy regularization, smoothness regularization, terminal position, and terminal attitude rewards is designed toward satisfying the multi-objective optimization requirements. In particular, the velocity-guidance terms encourage the vehicle to decelerate as it approaches the target, thereby reducing overshoot and oscillations.
\item A privileged multi-encoder architecture is proposed to enable the policy to respond rapidly to changes in ocean currents while generating smooth thrust commands. Ablation experiments validate the rationality and effectiveness of each module and further reveal that dynamic privileged information is the primary factor affecting action smoothness.
\item Simulation results demonstrate that the proposed TSRCA-PPO method consistently
outperforms the conventional cascaded P-PID controller across all evaluation
metrics. Specifically, TSRCA-PPO reduces the position error, attitude error,
settling time, energy index, and thrust-smoothness index
to \(42.7\%\), \(76.5\%\), \(10.6\%\), \(93.5\%\), and \(15.9\%\) of the
corresponding P-PID values, respectively.
\end{itemize}

The remainder of this paper is organized as follows. Section II introduces the problem formulation and underwater vehicle model. Section III introduces the environment, observation and privileged information. Section IV describes Methodology. Section V reports the ablation studies and comparative experiments. Section VI concludes the paper and discusses future work.

\section{Problem Formulation and Underwater Vehicle Model}
\subsection{Problem Formulation}
We consider the position-control task of an overactuated ROV equipped with eight thrusters in a time-varying underwater-current environment, where the vehicle’s internal parameters are uncertain and its observations are corrupted by noise. The goal of the position control task is to drive the vehicle toward a fixed target pose and maintain it there while minimizing energy consumption and suppressing thruster oscillations.

Given the current observation $\boldsymbol{o}_t\in\mathcal{O}$, the control policy outputs an action vector $\boldsymbol{a}_t$, where each element corresponds to one thruster command. The learning objective is therefore to find a feedback control law
\begin{equation}
	\pi_\theta : \mathcal{O} \rightarrow \boldsymbol{a}_t\in[-1,1]^8,
\end{equation}
that maximizes the expected discounted reward:
\begin{equation}
	\mathcal{J}(\theta)=
	\mathbb{E}_{\pi_\theta}\!\left[
	\sum_{t=0}^{T}\gamma^t r_t
	\right],
\end{equation}
where $r_t$ is the reward received at time step $t$, and $\gamma$ is the discount factor. The detailed design of the reward function is described in Section~\ref{sec:reward_function}. 
\subsection{Underwater Vehicle Model}
\begin{figure}[t]
	\centering
	\includegraphics[width=\linewidth]{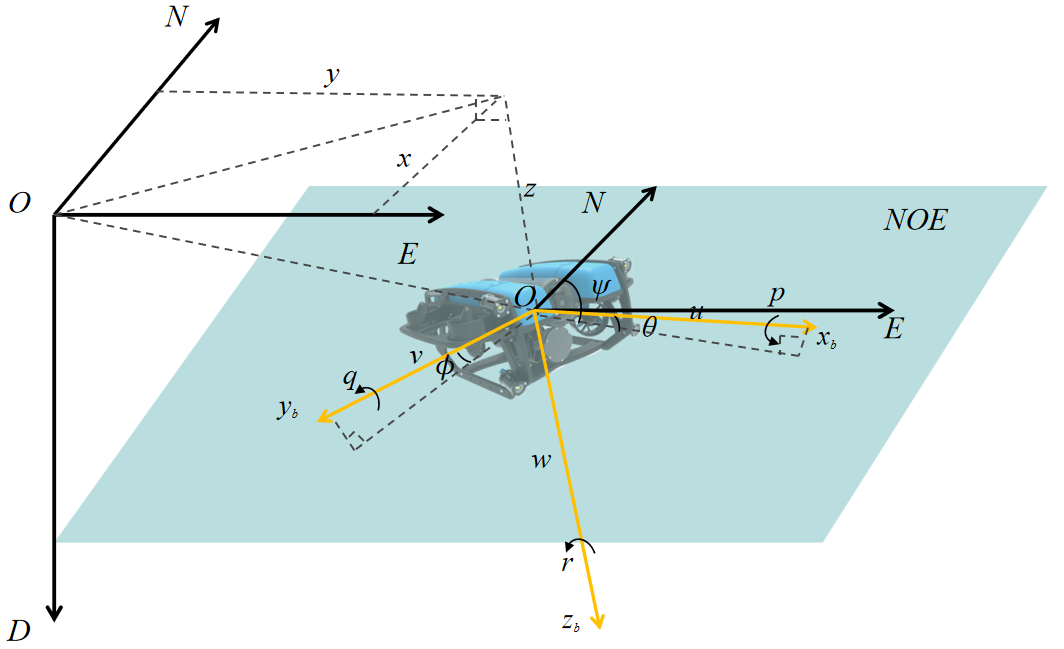}
	\caption{Coordinate transformation between the inertial frame and the body-fixed frame of the underwater vehicles.}
	\label{fig:coordinate_transformation}
\end{figure}
 In this paper, the North-East-Down (NED) inertial frame and body-fixed coordinate frame are adopted. The coordinate transformation between the two coordinate systems is shown in Fig.~\ref{fig:coordinate_transformation}. In the body coordinate, $x_b$ and $y_b$ point to the front and right of the vehicle, respectively, and $z_b$ is perpendicular to the $x_boy_b$ plane of the underwater vehicle. Let $\boldsymbol{\eta}=[x, y, z, \phi, \theta, \psi]^\top$ denote the position and Euler attitude of the vehicle in the inertial frame, and let $\boldsymbol{\nu}=[u, v, w, p, q, r]^\top$ denote the body-frame linear and angular velocities. A standard 6-DOF kinematic model of an underwater vehicle can be written as
\begin{equation}
	\dot{\boldsymbol{\eta}} = \boldsymbol{J}_{\Theta}(\boldsymbol{\eta})\boldsymbol{\nu},
	\label{eq:vehicle_kinematics}
\end{equation}
where $\boldsymbol{J}_{\Theta}(\boldsymbol{\eta})$ represents the Euler angular rotation matrix, and $\boldsymbol{\Theta}=[\phi,\theta,\psi]^\top$ represents the attitude of the vehicle. In addition, $c(\cdot)=\cos(\cdot)$, $s(\cdot)=\sin(\cdot)$, and $t(\cdot)=\tan(\cdot)$.
\begin{equation}
	\boldsymbol{J}_{\Theta}(\boldsymbol{\eta}) =
	\begin{bmatrix}
		\boldsymbol{J}_1(\boldsymbol{\Theta}) & \boldsymbol{0}_{3\times3}\\
		\boldsymbol{0}_{3\times3} & \boldsymbol{J}_2(\boldsymbol{\Theta})
	\end{bmatrix}
	\label{eq:jtheta_matrix}
\end{equation}
\begin{equation}
	\setlength{\arraycolsep}{2pt}
	\boldsymbol{J}_1(\boldsymbol{\Theta}) =
	\begin{bmatrix}
		c\psi c\theta & -s\psi c\phi + c\psi s\theta s\phi & s\psi s\phi + c\psi c\phi s\theta\\
		s\psi c\theta & c\psi c\phi + s\phi s\theta s\psi & -c\psi s\phi + s\theta s\psi c\phi\\
		-s\theta & c\theta s\phi & c\theta c\phi
	\end{bmatrix}
	\label{eq:j1_matrix}
\end{equation}
\begin{equation}
	\boldsymbol{J}_2(\boldsymbol{\Theta}) =
	\begin{bmatrix}
		1 & s\phi t\theta & c\phi t\theta\\
		0 & c\phi & -s\phi\\
		0 & s\phi/c\theta & c\phi/c\theta
	\end{bmatrix},\quad
	\theta \neq \frac{\pi}{2}
	\label{eq:j2_matrix}
\end{equation}

The vehicle dynamics model are given as follows\citep{dasilva2007modeling}:
\begin{equation}
	M\dot{\boldsymbol{\nu}}_r + C(\boldsymbol{\nu}_r)\boldsymbol{\nu}_r
	+ D(\boldsymbol{\nu}_r)\boldsymbol{\nu}_r + \boldsymbol{g}(\boldsymbol{\eta})
	= \boldsymbol{\tau}_c,
	\label{eq:vehicle_dynamics}
\end{equation}
where $\boldsymbol{\nu}_r$ is the velocity of the vehicle relative to the ocean current. $M\in\mathbb{R}^{6\times6}$ is the system inertia matrix, $C(\boldsymbol{\nu}_r)\in\mathbb{R}^{6\times6}$ is the Coriolis centripetal matrix, $D(\boldsymbol{\nu}_r)\in\mathbb{R}^{6\times6}$ represents the damping matrix, $\boldsymbol{g}(\boldsymbol{\eta})\in\mathbb{R}^{6}$ represents the restoring force vector, and $\boldsymbol{\tau}_c\in\mathbb{R}^{6}$ represents the control force vector. The system inertia matrix $M$ is expressed as:
\begin{equation}
	{\setlength{\arraycolsep}{4pt}
	M =
	\begin{bmatrix}
		m_u & 0 & 0 & 0 & mz_G & 0\\
		0 & m_v & 0 & -mz_G & 0 & 0\\
		0 & 0 & m_w & 0 & 0 & 0\\
		0 & -mz_G & 0 & I_p & 0 & 0\\
		mz_G & 0 & 0 & 0 & I_q & 0\\
		0 & 0 & 0 & 0 & 0 & I_r
	\end{bmatrix}}
	\label{eq:system_inertia_matrix}
\end{equation}
The Coriolis centripetal matrix $C(\boldsymbol{\nu}_r)$ is expressed as:
\begin{equation}
	C(\boldsymbol{\nu}_r)=
	\begin{bmatrix}
		\boldsymbol{0}_{3\times3} & C_{12}(\boldsymbol{\nu}_r)\\
		C_{21}(\boldsymbol{\nu}_r) & C_{22}(\boldsymbol{\nu}_r)
	\end{bmatrix},
	\label{eq:coriolis_block_matrix}
\end{equation}
where
\begin{equation}
	\resizebox{0.82\columnwidth}{!}{$
	C_{12}(\boldsymbol{\nu}_r)=
	\begin{bmatrix}
		mz_G r & m_w w_r & -m_v v_r\\
		-m_w w_r & mz_G r & m_u u_r\\
		-mz_G p + m_v v_r & -mz_G q - m_u u_r & 0
	\end{bmatrix}
	$}
	\label{eq:c12_matrix}
\end{equation}
\begin{equation}
	\resizebox{0.82\columnwidth}{!}{$
	C_{21}(\boldsymbol{\nu}_r)=
	\begin{bmatrix}
		-mz_G r & m_w w_r & mz_G p - m_v v_r\\
		-m_w w_r & -mz_G r & mz_G q + m_u u_r\\
		m_v v_r & -m_u u_r & 0
	\end{bmatrix}
	$}
	\label{eq:c21_matrix}
\end{equation}
\begin{equation}
	C_{22}(\boldsymbol{\nu}_r)=
	\begin{bmatrix}
		0 & I_r r & -I_q q\\
		-I_r r & 0 & I_p p\\
		I_q q & -I_p p & 0
	\end{bmatrix}
	\label{eq:c22_matrix}
\end{equation}
The symbols introduced in the matrices $M$ and $C(\boldsymbol{\nu}_r)$ are represented as follows:
\begin{equation}
	\begin{aligned}
		m_u &= m-X_{\dot{u}}, & I_p &= I_x-K_{\dot{p}},\\
		m_v &= m-Y_{\dot{v}}, & I_q &= I_y-M_{\dot{q}},\\
		m_w &= m-Z_{\dot{w}}, & I_r &= I_z-N_{\dot{r}}.
	\end{aligned}
	\label{eq:mass_inertia_symbols}
\end{equation}
At low speed, the hydrodynamic damping matrix $D(\boldsymbol{\nu}_r)=D+D_n(\boldsymbol{\nu}_r)$ is considered without coupling, so that the linear and nonlinear damping coefficients are contained in diagonal matrices~\citep{vonbenzon2022benchmark}.  The linear damping matrix is given by
\begin{equation}
	D=\operatorname{diag}(X_u,Y_v,Z_w,K_p,M_q,N_r),
	\label{eq:linear_damping_matrix}
\end{equation}
and the nonlinear damping matrix is
\begin{equation}
	\begin{aligned}
	D_n(\boldsymbol{\nu}_r)=\operatorname{diag}(&X_{u|u|}|u_r|,Y_{v|v|}|v_r|,Z_{w|w|}|w_r|,\\
	&K_{p|p|}|p|,M_{q|q|}|q|,N_{r|r|}|r|),
	\end{aligned}
	\label{eq:nonlinear_damping_matrix}
\end{equation}
where $D_n(\boldsymbol{\nu}_r)$ represents the quadratic damping matrix. The restoring force vector $\boldsymbol{g}(\boldsymbol{\eta})$, including gravity $W$ and buoyancy $B$, is expressed as:
\begin{equation}
	\boldsymbol{g}(\boldsymbol{\eta}) =
	\begin{bmatrix}
		(W-B)\sin\theta\\
		-(W-B)\cos\theta\sin\phi\\
		-(W-B)\cos\theta\cos\phi\\
		z_G W\cos\theta\sin\phi\\
		z_G W\sin\theta\\
		0
	\end{bmatrix}
	\label{eq:restoring_force_vector}
\end{equation}

An ROV such as the BlueROV2 Heavy is actuated by \(r=8\) fixed thrusters. The generalized control force generated by these thrusters is related to the thrust vector through
\begin{equation}
	\boldsymbol{\tau}_c=\boldsymbol{T}\boldsymbol{F}(t),
\end{equation}
where \(\boldsymbol{F}(t)=[F_1(t),\ldots,F_r(t)]^\top\) denotes the thrust vector produced by all thrusters at time $t$, and \(\boldsymbol{T}\in\mathbb{R}^{6\times r}\) is the thrust configuration matrix~\citep{fossen2002marine}:
\begin{equation}
	\boldsymbol{T}=\left[\boldsymbol{q}_1,\ldots,\boldsymbol{q}_r\right].
\end{equation}
Each column vector \(\boldsymbol{q}_i\) links the force generated by the \(i\)-th thruster to the generalized force and moment vector:
\begin{equation}
	\boldsymbol{q}_i=
	\begin{bmatrix}
		\boldsymbol{\epsilon}_i\\
		\boldsymbol{r}_i\times\boldsymbol{\epsilon}_i
	\end{bmatrix},
\end{equation}
where \(\boldsymbol{\epsilon}_i\) is a unit vector representing the orientation of thruster \(i\), and \(\boldsymbol{r}_i\) is the position vector from the body-frame origin to the point of application of the thrust \(F_i\).

The thruster model is constructed based on the T200 thruster and characterizes the mapping from throttle commands to generated thrust, accounting for the input dead zone, RPM saturation, actuation delay, and nonlinear thrust generation~\citep{chu2025marinegym}. 
Combining the vehicle dynamics in Eq.~\eqref{eq:vehicle_dynamics}, the thrust configuration relation, and the actuator model, the final throttle-to-motion equation is obtained as
\begin{equation}
	M\dot{\boldsymbol{\nu}}_r
	+ C(\boldsymbol{\nu}_r)\boldsymbol{\nu}_r
	+ D(\boldsymbol{\nu}_r)\boldsymbol{\nu}_r
	+ \boldsymbol{g}(\boldsymbol{\eta})
	= \boldsymbol{T}\boldsymbol{F}(t).
	\label{eq:final_throttle_to_motion}
\end{equation}

\section{Environment, Observation and Privileged Information}

\subsection{Simulation Environment}
\begin{figure}[!htbp]
	\centering
	\includegraphics[width=0.85\linewidth]{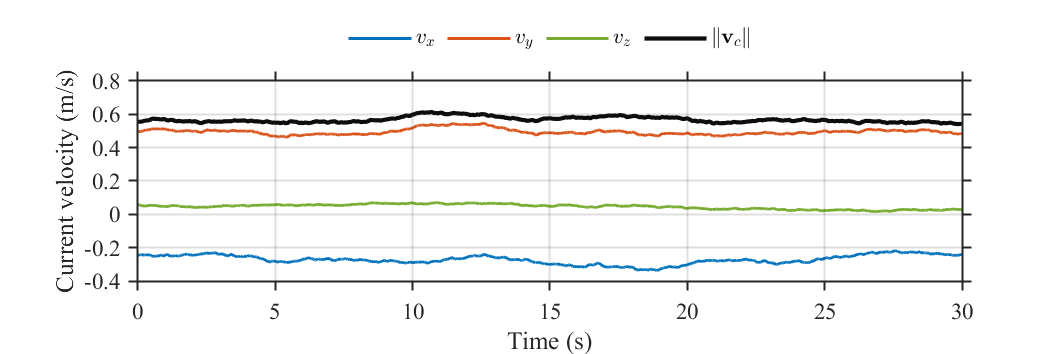}
	\caption{An example of a Gauss--Markov time-varying 3D ocean current over 30~s, with parameters \(V_c = 0.5494~\mathrm{m/s}\), \(\alpha = 7.45^\circ\), and \(\beta = 132.93^\circ\).}
	\label{fig:gauss_markov_flow}
\end{figure}

The simulation environment is built upon the Isaac-based framework introduced in~\citep{chu2025marinegym}, and the entire simulation process is discretized using a time step of $\Delta t=0.016~\mathrm{s}$. The nominal physical parameters of the vehicle are summarized in Table~\ref{tab:bluerovheavy_params}. In each simulation episode, the target position is fixed at $[0,0,0]^\top~\mathrm{m}$, while the target roll, pitch, and yaw angles are independently sampled from $[0,2\pi)$. The initial vehicle position is sampled from a cubic region with a side length of $4~\mathrm{m}$ centered at the origin, and its initial attitude angles are sampled from the same range as the target attitude. To avoid the singularities and discontinuities associated with Euler-angle representations and to facilitate policy optimization, both the target and vehicle attitudes are converted into rotation matrices before being provided to the controller~\citep{zhou2019continuity}. The current is assumed to be a bounded, irrotational, and slowly varying current and is represented through a first-order Gauss--Markov process~\citep{vu2021robust}. At the beginning of each episode, the mean current state
\(\bar{\boldsymbol{\xi}}=[\bar V_c,\bar\alpha,\bar\beta]^\top\) is sampled, where \(\bar V_c\), \(\bar\alpha\), and \(\bar\beta\) represent the mean speed magnitude, the vertical angle of attack, and the sideslip angle of the current, respectively. The instantaneous current state \(\boldsymbol{\xi}=[V_c,\alpha,\beta]^\top\) evolves as
\begin{equation}
\boldsymbol{\xi}_{t+1}
=\boldsymbol{\xi}_{t}
-\boldsymbol{a}_{c}\odot(\boldsymbol{\xi}_{t}-\bar{\boldsymbol{\xi}})\Delta t
+\boldsymbol{\sigma}_{c}\odot\sqrt{\Delta t}\,\boldsymbol{\epsilon}_{t},
\end{equation}
where \(\boldsymbol{\epsilon}_{t}\sim\mathcal{N}(\boldsymbol{0},I)\), \(\boldsymbol{a}_c\) denotes the mean-reversion rate of the current state toward the episode-level mean \(\bar{\boldsymbol{\xi}}\), while \(\boldsymbol{\sigma}_c\) represents the stochastic perturbation intensity of the Gauss--Markov current process. The resulting world-frame current velocity is
\begin{equation}
\boldsymbol{u}_{\mathrm{flow}}
=
\begin{bmatrix}
V_c\cos\alpha\cos\beta\\
V_c\sin\beta\\
V_c\sin\alpha\cos\beta
\end{bmatrix}.
\end{equation}
The fixed Gauss--Markov current-model parameters, which capture the slowly varying characteristics of ocean currents, are listed in Table~\ref{tab:bluerovheavy_params}, while the randomized current parameters, which govern the magnitude and direction of the ocean current, are summarized in Table~\ref{tab:dr_summary}.  Fig.~\ref{fig:gauss_markov_flow} displays an example of a Gauss--Markov time-varying 3D ocean current over 30~s, which slowly fluctuates around its episode-level mean.

To simulate realistic sensing conditions, observation noise is introduced into the environment. The noise is modeled as zero-mean Gaussian noise, and the corresponding standard deviations are listed in Table~\ref{tab:obs_noise_std}. To adapt to diverse model uncertainties of vehicles, domain randomization is applied at every episode reset. As reported in \citet{vonbenzon2022benchmark}, the rotational added-mass terms are subject to relatively large modeling errors. Therefore, a wider domain-randomization range of $[0.5,1.5]$ is applied to the rotational added-mass coefficients,  while the remaining randomized parameters and sampling ranges are summarized in Table~\ref{tab:dr_summary}.

\begin{table}[t]
\centering
\caption{Nominal vehicle and fixed current-model parameters used in simulation.}
\label{tab:bluerovheavy_params}
\setlength{\tabcolsep}{4pt}
\renewcommand{\arraystretch}{1.1}
\resizebox{\linewidth}{!}{%
\begin{tabular}{ll}
\hline
Parameter & Value \\
\hline
Mass $m$ (kg) & $11.2$ \\
Volume $V$ (m$^3$) & $0.0113459$ \\
Center-of-buoyancy offset $c_{oB}$ (m) & $0.01$ \\
Added-mass diagonal $\boldsymbol{M}_A$ & $\mathrm{diag}(5.5,\ 12.7,\ 14.57,\ 0.12,\ 0.12,\ 0.12)$ \\
Linear damping diagonal $\boldsymbol{D}_L$ & $\mathrm{diag}(4.03,\ 6.22,\ 5.18,\ 0.07,\ 0.07,\ 0.07)$ \\
Quadratic damping diagonal $\boldsymbol{D}_Q$ & $\mathrm{diag}(18.18,\ 21.66,\ 36.99,\ 1.55,\ 1.55,\ 1.55)$ \\
Inertia diagonal $\boldsymbol{I}_0$ (kg\,m$^2$) & $\mathrm{diag}(0.30375,\ 0.62600,\ 0.57690)$ \\
Number of rotors $N_r$& $8$ \\
Motor force constant $k_F$ & $1.0$ \\
Thruster delay time constant $T_d$ (s) & $0.1$ \\
Gauss--Markov rate $\boldsymbol{a}_c$ (s$^{-1}$) & $[1/8,\ 1/10,\ 1/12]^\top$ \\
Gauss--Markov noise $\boldsymbol{\sigma}_c$ & $[0.035~\mathrm{m/s},\ 1.80^\circ,\ 4.00^\circ]^\top$ \\
\hline
\end{tabular}
}
\end{table}

\subsection{Observation and Privileged Information}
 The observation space is defined using variables that can be measured or estimated by onboard sensors, with all spatial vectors consistently expressed in the inertial frame and the attitude represented by the body-to-inertial rotation matrix. Position, orientation, linear velocity, angular velocity, and linear acceleration are assumed to be directly measurable using onboard sensors, and their assumed measurement accuracies are summarized in Table~\ref{tab:obs_noise_std}. A single-frame actor observation is constructed as
 \begin{equation}
 	\tilde{\boldsymbol{o}}_t =
 	\left[
 	\hat{\boldsymbol{r}}_{p,t},\
 	\operatorname{vec}(\hat{\boldsymbol{R}}_t),\
 	\operatorname{vec}(\boldsymbol{R}_t^d),\
 	\hat{\boldsymbol{v}}_t,\
 	\hat{\boldsymbol{\omega}}_t,\
 	\tilde{\dot{\boldsymbol{v}}}_t
 	\right],
 \end{equation}
 where the relative position error vector $\hat{\boldsymbol{r}}_{p,t}=\boldsymbol{p}_t^d-\hat{\boldsymbol{p}}_t$, linear velocity $\hat{\boldsymbol{v}}_t$, attitude rotation matrix $\hat{\boldsymbol{R}}_t$, and angular velocity $\hat{\boldsymbol{\omega}}_t$ are optimized by the extended Kalman filter (EKF).
 Linear acceleration $\tilde{\dot{\boldsymbol{v}}}_t$ is modeled as a noisy onboard measurement.  $\boldsymbol{R}_t^d$ is the desired attitude rotation matrix, and $\operatorname{vec}(\cdot)$ denotes column-wise vectorization of a matrix. 
 In addition to these directly measurable quantities, information that is either unobservable or costly to obtain such as ocean-current conditions, angular acceleration, the actual motor rotational speeds, and the vehicle’s true physical parameters can also improve control performance. In the following, we estimate these quantities using the learning-by-cheating paradigm~\citep{chen2020learning}, where a teacher policy is first trained using privileged global information, after which an implicit representation of the privileged state is distilled into a student policy that relies only on observations available during deployment.

Privileged information is defined as variables that are available in simulation but cannot be directly measured or reliably estimated by the vehicle's onboard sensors during real-world deployment, and we summarize its composition as
\begin{equation}
\boldsymbol{I}^{priv}_t=
\left[
\begin{aligned}
&\underbrace{m,\ c_{oB},\ \boldsymbol{I}_0,\ V,\ \boldsymbol{k}_F,\ \boldsymbol{T}_d,\ \boldsymbol{M}_A,\ \boldsymbol{D}_L,\ \boldsymbol{D}_Q}_{\boldsymbol{S}_t:\,\text{True Static Parameters}},\\
&\underbrace{\boldsymbol{o}_t,\ \boldsymbol{\mathrm{rpm}}_t}_{\boldsymbol{D}_t:\,\text{True Dynamic States}},\quad \underbrace{\boldsymbol{\xi}_{t}}_{\boldsymbol{\xi}_t:\,\text{True Current Parameters}}
\end{aligned}
\right],
\end{equation}
where the privileged vector is composed of three groups of information.
The three groups are denoted by $\boldsymbol{S}_t$, $\boldsymbol{D}_t$, and $\boldsymbol{\xi}_t$, representing the true static parameters, the true dynamic states comprising the onboard state and actual motor rotational speeds, and the true current parameters, respectively. Specifically, $\boldsymbol{S}_t$ contains the vehicle mass $m$, center-of-buoyancy offset $c_{oB}$, inertia matrix $\boldsymbol{I}_0$, displaced volume $V$, motor force constants $\boldsymbol{k}_F$, thruster delay time constants $\boldsymbol{T}_d$, added-mass matrix $\boldsymbol{M}_A$, and linear and quadratic damping matrices $\boldsymbol{D}_L$ and $\boldsymbol{D}_Q$. The dynamic group $\boldsymbol{D}_t=[\boldsymbol{o}_t,\boldsymbol{\mathrm{rpm}}_t]$ consists of the noise-free onboard state $\boldsymbol{o}_t$, including the relative position error, current and desired attitude rotation matrices, linear velocity, angular velocity, and linear acceleration, together with the actual rotational speeds $\boldsymbol{\mathrm{rpm}}_t$ of the eight thrusters. With access to these privileged variables, the teacher policy can obtain a more complete description of the current vehicle--environment interaction, which facilitates learning a high-quality action mapping during training.

\begin{table}[t]
\centering
\caption{Domain randomization ranges used in simulation.}
\label{tab:dr_summary}
\setlength{\tabcolsep}{4pt}
\renewcommand{\arraystretch}{1.1}
\begin{tabular}{p{0.50\linewidth}p{0.40\linewidth}}
\hline
Randomized term & Range/value \\
\hline
Body mass scale & $[0.8,1.2]$ \\
Inertia scale & $[0.8,1.2]$ \\
Volume scale & $[0.8,1.2]$ \\
Center-of-buoyancy offset scale & $[-3.0,3.0]$ \\
Translational added-mass scale & $[0.8,1.2]$ \\
Rotational added-mass scale & $[0.5,1.5]$ \\
Linear damping scale & $[0.8,1.2]$ \\
Quadratic damping scale & $[0.8,1.2]$ \\
Rotor force-constant scale & $[0.8,1.2]$ \\
Mean current speed $\bar V_c$ (m/s) & $[0.20,0.60]$ \\
Mean vertical current angle $\bar\alpha$ & $[-8^\circ,8^\circ]$ \\
Mean horizontal current angle $\bar\beta$ & $[-180^\circ,180^\circ]$ \\
\hline
\end{tabular}
\end{table}

\begin{table}[t]
\centering
\caption{Observation noise standard deviations used in simulation.}
\label{tab:obs_noise_std}
\setlength{\tabcolsep}{4pt}
\renewcommand{\arraystretch}{1.1}
\footnotesize
\begin{tabular}{ll}
\hline
Noise term & Standard deviation \\
\hline
Position $\boldsymbol{r}_p$ & $0.02~\mathrm{m}$ \\
Attitude $\boldsymbol{\eta}_{\mathrm{attitude}}$ & $0.03~\mathrm{rad}$ \\
Linear velocity $\boldsymbol{v}$ & $0.01~\mathrm{m/s}$ \\
Angular velocity $\boldsymbol{\omega}$ & $0.02~\mathrm{rad/s}$ \\
Linear acceleration $\dot{\boldsymbol{v}}$ & $0.05~\mathrm{m/s^2}$  \\
\hline
\end{tabular}
\end{table}

\section{Methodology}
\subsection{Overall Framework: Two-stage Training Pipeline of TSRCA-PPO}
\begin{figure}[t]
	\centering
	\includegraphics[width=\linewidth]{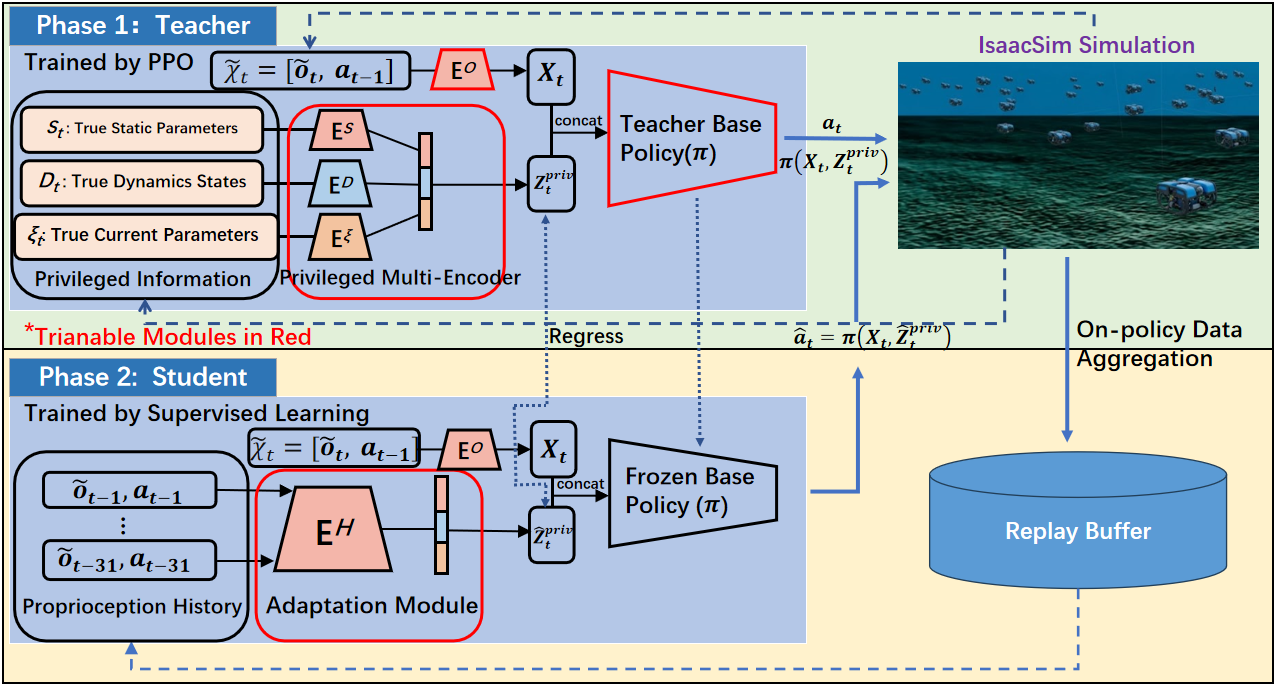}
	\caption{Overall framework of the two-stage training pipeline of TSRCA-PPO.}
	\label{fig:overall_framework}
\end{figure}

The complete framework is shown in Fig.~\ref{fig:overall_framework}, which follows a two-stage training pipeline. In Stage I, a teacher policy is trained using PPO with privileged information containing $\boldsymbol{S}_t$ (True Static Parameters), $\boldsymbol{D}_t$ (True Dynamic States), and $\boldsymbol{\xi}_t$ (True Current Parameters). The privileged-information multi-encoder comprises three encoders, $E^{S}$, $E^{D}$, and $E^{\xi}$, whose outputs are concatenated to form the final latent privileged feature $\boldsymbol{z}_t^{priv}$. The current action--observation vector $\tilde{\boldsymbol{\chi}}_t$ is mapped by the action--observation encoder $E^{O}$ to the latent action--observation feature $\boldsymbol{x}_t$, which is concatenated with the latent privileged feature $\boldsymbol{z}_t^{priv}$ and then fed into the teacher base-policy module $\pi$. The multi-encoder and base policy are then jointly optimized to generate the optimal action.  In Stage II, the base policy is frozen, and an adaptation module is trained via supervised learning to distill the privileged-information multi-encoder. The estimated latent privileged feature $\hat{\boldsymbol{z}}_t^{priv}$ is concatenated with the latent action--observation feature $\boldsymbol{x}_t$ and then fed into the frozen base policy, thereby yielding the student policy. Specifically, data collection is performed using Dataset Aggregation (DAgger)~\citep{ross2011dagger}, and the estimated latent privileged feature $\hat{\boldsymbol{z}}_t^{priv}$ is estimated from proprioception history.
\subsection{Stage I: Teacher Policy Training under Privileged PPO Framework}
\begin{figure}[t]
	\centering
	\includegraphics[width=\linewidth]{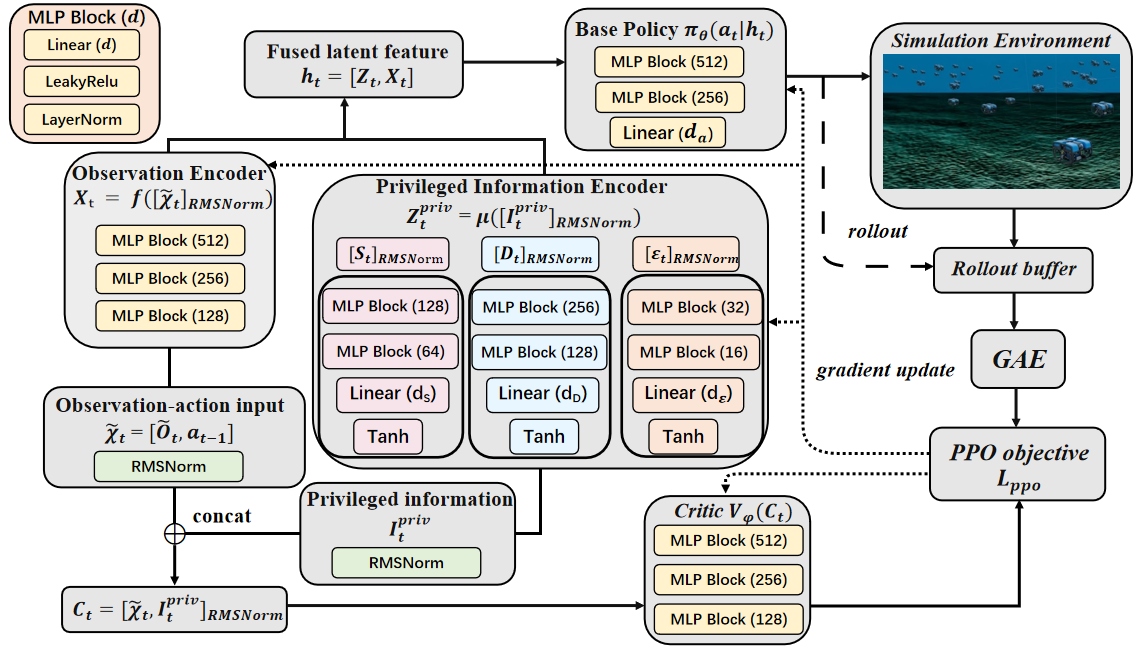}
	\caption{Stage-I privileged PPO framework.}
	\label{fig:ppo_actor_critic}
\end{figure}

As illustrated in Fig.~\ref{fig:ppo_actor_critic}, a teacher policy is trained in Stage~I using PPO with privileged information \citep{schulman2017proximal}, where the teacher has access to the global state information to generate optimal actions. Concretely, for the actor, the action--observation input
$\tilde{\boldsymbol{\chi}}_t
=
[\tilde{\boldsymbol{o}}_t,\boldsymbol{a}_{t-1}]$
is first encoded by an observation encoder. Meanwhile, the three components
of the privileged information, namely the true static parameters
$\boldsymbol{S}_t$, true dynamic states $\boldsymbol{D}_t$, and true current
parameters $\boldsymbol{\xi}_t$, are independently compressed into their
respective latent spaces by three dedicated encoders. The resulting latent
representations are then concatenated with the action--observation embedding
and fed into the base policy network, which ultimately outputs the action vector $\boldsymbol{a}_t$. For the critic, it directly receives the concatenated input $\boldsymbol{c}_t=[\tilde{\boldsymbol{\chi}}_t,\boldsymbol{I}_t^{\mathrm{priv}}]$. We adopt a low-dimensional implicit actor latent space because high-dimensional representations make the student policy more difficult to learn and tend to increase regression errors. Although dimensionality reduction inevitably discards some information, the resulting compact representation retains sufficient task-relevant information for control. A standard PPO training procedure is then executed to optimize the actor and critic networks, where the advantages value function of critic is estimated through generalized advantage estimation (GAE)~\citep{schulman2016gae},
\begin{equation}
	\hat{A}_t=\sum_{l=0}^{\infty}(\gamma\lambda)^l\delta^V_{t+l},
	\quad
	\delta^V_t=r^{\mathrm{env}}_t+\gamma V_\varphi(\boldsymbol{c}_{t+1})-V_\varphi(\boldsymbol{c}_t),
\end{equation}
and the actor is optimized using the clipped surrogate objective
\begin{equation}
	\mathcal{L}_{\mathrm{clip}}
	=-\mathbb{E}_t\!\left[
	\min\!\left(
	\rho_t(\theta)\hat{A}_t,\,
	\mathrm{clip}(\rho_t(\theta),1-\epsilon,1+\epsilon)\hat{A}_t
	\right)
	\right],
\end{equation}
This clipping mechanism constrains the deviation between the updated and previous policies, thereby preventing excessively large policy updates and improving training stability. The training loss is finally defined as
\begin{equation}
	\mathcal{L}_{\mathrm{PPO}}=
	\mathcal{L}_{\mathrm{clip}}
	+0.5\,\mathcal{L}_{V}
	+\lambda_{\mathrm{ent}}\mathcal{L}_{\mathrm{ent}}
	,
\end{equation}
where $\mathcal{L}_{V}$ is the value loss and $\mathcal{L}_{ent}$ is the entropy loss. $\mathcal{L}_{V}$ trains the value function to provide accurate advantage estimates, while $\mathcal{L}_{ent}$ encourages policy exploration by penalizing overly deterministic action distributions.

The training implementation is built upon the CleanRL codebase~\citep{huang2022cleanrl}. We use the Adam optimizer, and all network components in the proposed framework are implemented as lightweight multilayer perceptrons. All latent features are mapped to the range $[-1,1]$ using the Tanh activation, which standardizes the numerical scales of the static, state, and flow latent branches and improves the stability of reinforcement learning.

\subsection{Stage II: Student Policy Training}
\begin{figure}[t]
\centering
\includegraphics[width=\linewidth]{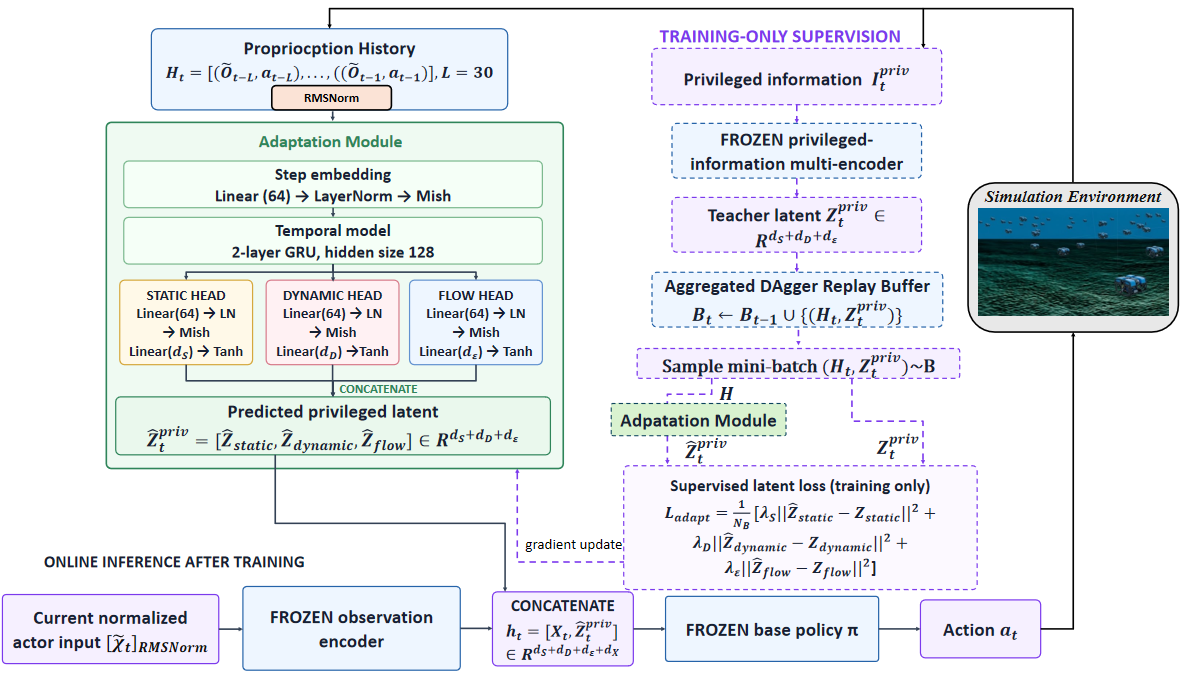}
\caption{Stage-II adaptation module architecture.}
\label{fig:adaptation_module}
\end{figure}

Fig.~\ref{fig:adaptation_module} shows the Stage-II adaptation module architecture. In Stage~II, the base policy and the observation encoder is frozen. The teacher policy trained in Stage~I is used to generate true-state latent features for training the adaptation module to replace the privileged-information multi-encoder through supervised learning. To reduce the distribution shift between training and deployment, we adopt a DAgger-style on-policy data aggregation strategy. Specifically, the deployable student policy is used to interact with the environment and generate trajectories. The resulting onboard observations and executed actions are then appended to a history buffer:
\begin{equation}
	\mathbf{H}_t = \left[(\tilde{\mathbf{o}}_{t-L},\mathbf{a}_{t-L}),\ldots,(\tilde{\mathbf{o}}_{t-1},\mathbf{a}_{t-1})\right],
\end{equation}
where $L$ is the history length. The privileged information multi-encoder provides the teacher latent code $\mathbf{z}_t^{priv}$ from privileged intrinsics, while the adaptation module predicts $\hat{\mathbf{z}}_t^{priv}$ from $\mathbf{H}_t$. The replay buffer stores paired samples of the proprioceptive history
$\mathbf{H}_t$ and the corresponding privileged teacher latent
$\mathbf{z}_t^{\mathrm{priv}}$ and is maintained at a sufficiently large capacity to ensure stable random sampling during data updates. The training of the adaptation module is performed concurrently with data collection. After each data-collection phase, the replay buffer is updated in a first-in-first-out manner. The loss function is described as:
 \begin{equation}
 \begin{aligned}
 \mathcal{L}_{\mathrm{adapt}}
 ={}& \frac{1}{N_B}\Bigg[\lambda_S\left\|\hat{\mathbf{z}}_{\mathrm{static}}-\mathbf{z}_{\mathrm{static}}\right\|_2^2 \\
 &+\lambda_D\left\|\hat{\mathbf{z}}_{\mathrm{dynamic}}-\mathbf{z}_{\mathrm{dynamic}}\right\|_2^2 \\
 &+\lambda_{\epsilon}\left\|\hat{\mathbf{z}}_{\mathrm{flow}}-\mathbf{z}_{\mathrm{flow}}\right\|_2^2\Bigg].
 \end{aligned}
 \end{equation}
 where $N_B$ denotes the mini-batch size, and $\lambda_S$, $\lambda_D$, and $\lambda_{\epsilon}$ are the weighting coefficients for the static-parameter, dynamic-state, and current-flow latent-feature losses, respectively.

The adaptation module first embeds the history information $\mathbf{H}_t$ and then uses a two-layer GRU to extract temporal features. The resulting shared temporal representation is subsequently fed into three branch-specific prediction heads to estimate the static, dynamic, and flow latent features, which are finally concatenated to form the complete latent representation. Mish activations are employed in the Phase-II adaptation module to provide smooth nonlinear transformations for temporal latent regression, while preserving informative negative responses and improving optimization stability.

\subsection{Reward Function}
\label{sec:reward_function}
In this work, the reward function at time step t is composed of two parts: a terminal exponential-shaped term and a dense process-guided term:
\begin{equation}
	r_t=r_{\mathrm{term,t}}+r_{\mathrm{proc,t}}.
\end{equation}
For the terminal term part, let $e_{p,t}=p^d-p_{t}$ denote the position-error vector, $\boldsymbol{q_t}$ and $\boldsymbol{q}_d$ denote current/target quaternions, $\boldsymbol{v}$ and $\boldsymbol{\omega}$ be linear and angular velocities. Define the quaternion error:
\begin{equation}
	\boldsymbol{q}_{e,t}=\boldsymbol{q}_d^{-1}\otimes \boldsymbol{q_t}
	=\begin{bmatrix}q_{e,w,t}\\ \boldsymbol{q}_{e,v,t}\end{bmatrix},
\end{equation}
and convert quaternion error to degree errors as
\begin{equation}
	e_{q,t}^{\circ}=\frac{180}{\pi}\,2\arccos\!\left(\left|\langle \boldsymbol{q_t},\boldsymbol{q}_d\rangle\right|\right).
\end{equation}
The terminal reward at the $t$-th time step is defined as:
\begin{equation}
	r_{\mathrm{term,t}}=5d_{p,t}+3d_{q,t}.
\end{equation}
where
\begin{equation}
	d_{p,t}=e^{-20e_{p,t}},\quad
	d_{q,t}=e^{-0.2e_{q,t}^{\circ}}.
\end{equation}
\begin{figure}[t]
	\centering
	\captionsetup[subfloat]{font=scriptsize}
	\subfloat[Terminal position reward term.\label{fig:reward_dp}]{%
		\begin{tikzpicture}
			\begin{axis}[
				width=0.49\linewidth,
				height=0.32\linewidth,
				xlabel={$e_p$ (m)},
				ylabel={$d_p$},
				xmin=0, xmax=0.30,
				ymin=0, ymax=1.05,
				xtick={0,0.10,0.20,0.30},
				xticklabels={0,0.1,0.2,0.3},
				grid=both,
				major grid style={black!15},
				minor grid style={black!7},
				tick label style={font=\scriptsize},
				label style={font=\scriptsize},
				]
				\addplot[
				blue,
				thick,
				domain=0:0.30,
				samples=200
				] {exp(-20*x)};
			\end{axis}
		\end{tikzpicture}
	}
	\hfill
	\subfloat[Terminal attitude reward term.\label{fig:reward_dq}]{%
		\begin{tikzpicture}
			\begin{axis}[
				width=0.49\linewidth,
				height=0.32\linewidth,
				xlabel={$e_q^\circ$ (deg)},
				ylabel={$d_q$},
				xmin=0, xmax=30,
				ymin=0, ymax=1.05,
				grid=both,
				major grid style={black!15},
				minor grid style={black!7},
				tick label style={font=\scriptsize},
				label style={font=\scriptsize},
				]
				\addplot[red!75!black, thick, domain=0:30, samples=160] {exp(-0.2*x)};
			\end{axis}
		\end{tikzpicture}
	}
	\caption{Terminal exponential reward terms for position and attitude errors.}
	\label{fig:reward_terminal_factors}
\end{figure}
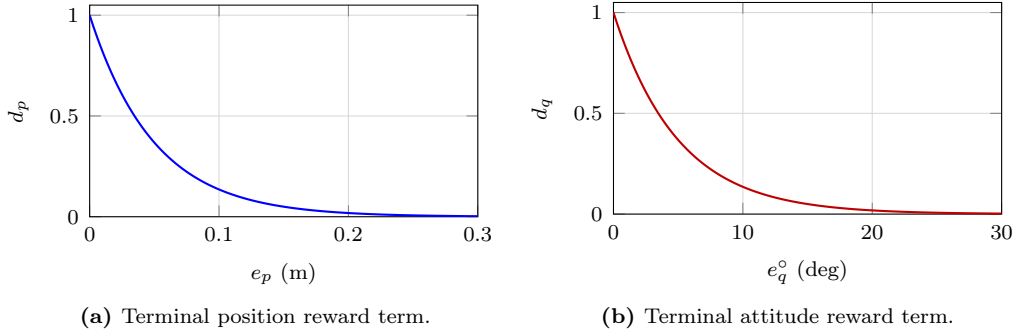
As shown in Fig.~\ref{fig:reward_terminal_factors}, the terminal exponential reward terms become effective when the position and attitude errors are small, and increase rapidly as the errors approach zero. This design assigns the maximum reward when the vehicle remains at the target pose, encouraging it to approach the target and maintain stabilization near the desired hovering state.

For the process term part, the position guidance reward is defined using the projection of the vehicle velocity onto the position-error vector:
\begin{equation}
	r_{proc,p,t}=4.5\left\langle \boldsymbol{v_t},\frac{\boldsymbol{e}_{p,t}}{\|\boldsymbol{e}_{p,t}\|+\epsilon}\right\rangle,
\end{equation}
Similarly, the orientation guidance reward is defined as the signed scalar projection of the vehicle's angular velocity onto the quaternion-error vector:
\begin{equation}
	r_{proc,q,t}=2\langle\boldsymbol{\omega_t},\operatorname{sgn}(q_{e,w,t})\,\boldsymbol{q}_{e,v,t}\rangle.
\end{equation}
Both guidance terms are formulated as inner products between the desired correction directions and the corresponding velocity vectors. The translational term encourages the vehicle to move toward the target position, while the rotational term encourages the vehicle to rotate toward the desired attitude along the shortest rotation axis. When the vehicle approaches the target position or attitude, excessive linear or angular velocity is likely to cause overshoot, resulting in motion opposite to the subsequent correction direction and thus a larger negative reward. Consequently, the proposed velocity-guidance reward naturally encourages timely deceleration near the target, thereby mitigating overshoot and oscillations. Their coefficients are set smaller than those of the terminal reward terms, allowing the guidance rewards to accelerate convergence without dominating the final pose-control objective. 

Propulsion power is approximated by a cubic RPM proxy. This approximation is motivated by the standard open-water propeller model, where the propeller torque coefficient satisfies $ K_Q=Q_p/(4\pi^2\rho |\omega|\omega D^5)$, and the consumed propeller power is $P_p=\omega Q_p$. Therefore, when the variation of $K_Q$ is limited under low-speed station-keeping conditions, the propulsion power magnitude can be approximated as proportional to $|\omega|^3$\citep{pivano2008four}:
\begin{equation}
	E_t=\sum_{i=1}^{8}\left|\frac{n_{i,t}}{60}\right|^3.
	\label{eq:energy_proxy}
\end{equation} Accordingly, Eq.~\eqref{eq:energy_proxy} can be used as an approximate proxy for energy consumption: a larger $E_t$ indicates higher energy consumption, and vice versa. The reward is formulated as an exponential function to reduce the energy-consumption proxy during motion:
\begin{equation}
	r_{e,t}=12\exp(-2.5\times10^{-6}E_t).
\end{equation}
where $n_{i,t}$ denotes the actual rotational speed of the $i$-th motor at time step $t$. Smoothness reward is defined as
\begin{equation}
	\label{eq:force_smooth_reward}
	r_{s,t}=\exp\!\left(-\left\|\boldsymbol{a}_t-\boldsymbol{a}_{t-1}\right\|_1\right).
\end{equation}
The smoothness reward penalizes large changes between consecutive action vectors, thereby encouraging smoother control commands. The process reward is then summarized as
\begin{equation}
	r_{\mathrm{proc}}=r_{proc,p}+r_{proc,q}+r_e+r_s.
\end{equation}

\section{Ablation Studies and Comparative Experiments}
\subsection{Ablation Studies}
\begin{figure}[H]
	\centering
	\includegraphics[width=0.9\linewidth]{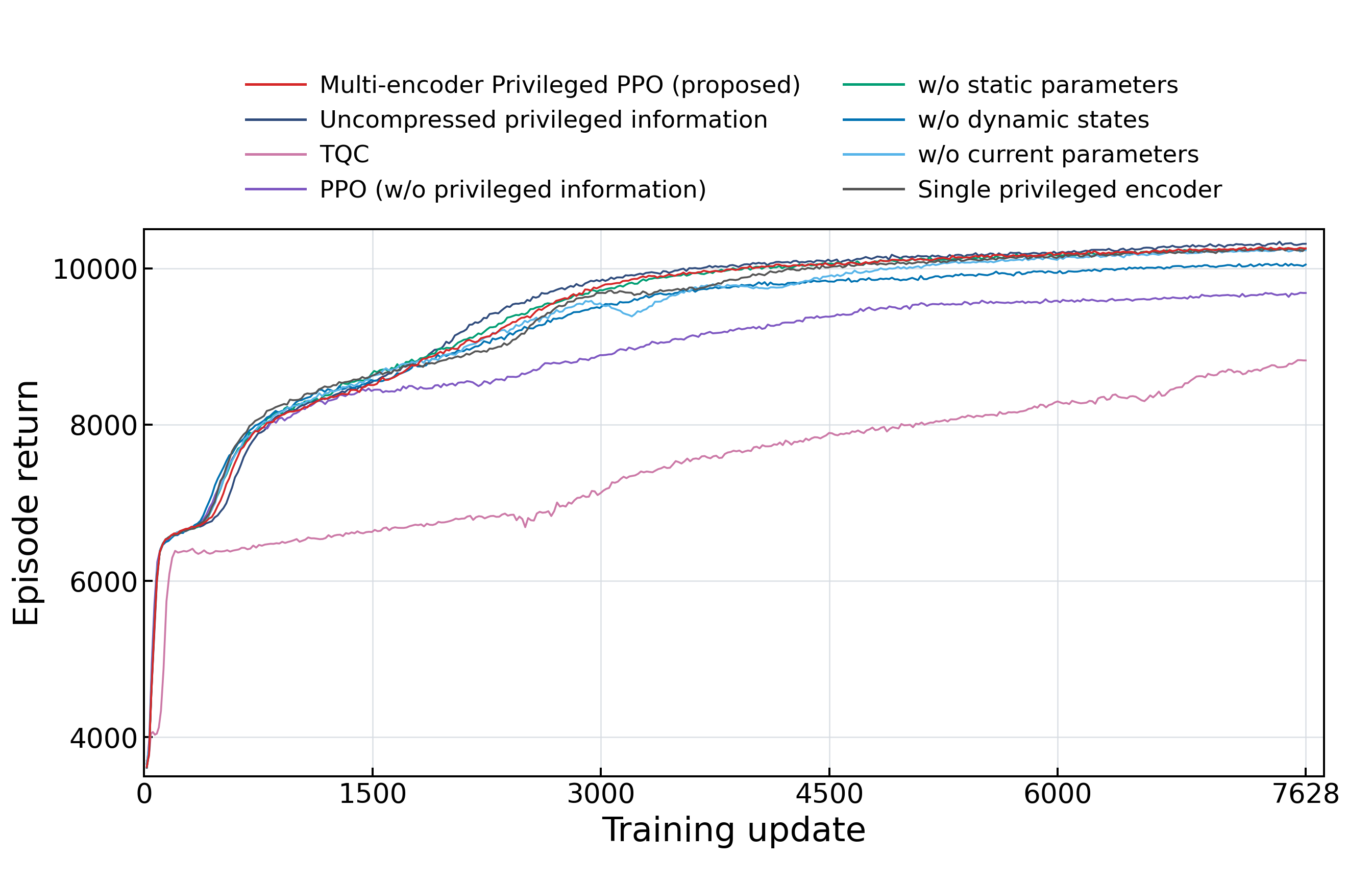}
	\caption{Stage-I training return curves for the proposed Privileged PPO architecture and seven ablation or baseline variants.}
	\label{fig:ablation_training_returns}
\end{figure}

Fig.~\ref{fig:ablation_training_returns} shows the Stage-I ablation training return curves.  In all ablation studies, only the model architecture is modified, while the reward function, observation space, and all other experimental settings remain unchanged For Stage-I training, we compare the proposed Privileged PPO architecture, the uncompressed privileged-information variant, a single privileged encoder, three variants that omit the static, dynamic, or current privileged-information group, PPO, and TQC. Removing privileged information leads to a substantial drop in return, whereas removing dynamic information causes only a modest decrease. Both the single-encoder architecture and the removal of flow information result in unstable training, while static information has relatively little impact during Stage I. The uncompressed variant, whose latent dimension is equal to the dimension of the privileged-information space, achieves the highest return, likely benefiting from its larger parameter capacity. However, the resulting high-dimensional latent representation also increases the difficulty of Stage-II adaptation and distillation. We therefore conclude that, during Stage-I training, PPO outperforms TQC, privileged information is essential, and dynamic-state information is the most important privileged-information component. 

 Before presenting the quantitative analysis, we first define the evaluation metrics considered in this study, including the position steady-state error, attitude steady-state error, settling time, success rate, energy consumption proxy, force smoothness, and inference latency. The steady-state errors are defined as the mean errors over the final one second of the evaluation horizon. With
\begin{equation}
	\mathcal{S}=\{k\mid T_{\mathrm{eval}}-1\le t_k\le T_{\mathrm{eval}}\},
\end{equation}
and the position and attitude steady-state error as:
\begin{equation}
	\mathrm{SS}_{\mathrm{pos}}=\frac{1}{|\mathcal{S}|}\sum_{t\in\mathcal{S}}e_{p,t},\qquad
	\mathrm{SS}_{\mathrm{att}}=\frac{1}{|\mathcal{S}|}\sum_{t\in\mathcal{S}}e_{q,t}^{\circ}.
\end{equation}
The settling time $t_s$ is the earliest time after which the position and attitude errors remain within the prescribed thresholds, and the success rate is the percentage of evaluation episodes that satisfy this criterion. In this study, the prescribed thresholds are set to $0.02$~m for the position error and $2^\circ$ for the attitude error and the settling time is computed only over successful episodes. Inference latency denotes the wall-clock time required for one policy step.
The energy consumption proxy is obtained by summing the per-step propulsion-energy proxy over time:
\begin{equation}
	J_E=\sum_{t=1}^{N} E_t
	=\sum_{t=1}^{N}\sum_{i=1}^{8}\left|\frac{n_i(t)}{60}\right|^3 .
\end{equation}
As described in Section~\ref{sec:reward_function}, the per-step energy proxy $E_t$ is approximately proportional to propulsion power under low-speed station-keeping conditions; consequently, its accumulated value $J_E$ can be used to approximate the relative magnitude of the actual propulsion energy consumption.
The force smoothness is quantified by the mean magnitude of the second-order force difference over the evaluation horizon:
\begin{equation}
	\begin{aligned}
		J_{\mathrm{FS}}
		&=\frac{1}{N_r(N-2)}\sum_{t=2}^{N-1}
		\left\|\boldsymbol{F}_{t+1}+\boldsymbol{F}_{t-1}-2\boldsymbol{F}_{t}\right\|_1,
	\end{aligned}
	\label{eq:force_smoothness_metric}
\end{equation}
where $N_r$ is the number of motors and $\boldsymbol{F}_t=[F_1(t),\ldots,F_{N_r}(t)]^\top$ is the actual thruster-force vector at the $t$-th instant. Here, we do not use the conventional sum of absolute action differences as the smoothness metric, because action commands are subject to actuator delay and therefore do not directly represent the actual motor response. Instead, we evaluate force smoothness, which more directly reflects variations in motor rotational speed and the resulting actuator output. A smaller $J_{\mathrm{FS}}$ indicates smoother force variation.
The latent-feature estimation accuracy is quantified by the $Z$ mean squared error (MSE):
\begin{equation}
	\mathrm{MSE}_{Z}=\frac{1}{N d_z}\sum_{t=1}^{N}
	\left\|\hat{\boldsymbol{z}}_t^{\mathrm{priv}}-\boldsymbol{z}_t^{\mathrm{priv}}\right\|_2^2,
	\label{eq:z_mse_metric}
\end{equation}
where $d_z$ denotes the dimension of the privileged latent feature and $N$ denotes the trajectory length measured in time steps. A smaller $\mathrm{MSE}_{Z}$ indicates more accurate reconstruction of the teacher latent feature by the adaptation module. 

Table~\ref{tab:stage1_ablation_metrics} presents the Stage-I experimental results in a 32-s station-keeping test. The results show that the PPO baseline can still achieve high steady-state accuracy and reaches the prescribed position and attitude error thresholds in 87.6\% of the evaluation trials without privileged information, supporting the effectiveness of the proposed reward design. However, its settling time increases substantially and force smoothness deteriorates markedly, while the steady-state position and attitude errors increase slightly. TQC exhibits a relatively large standard deviation in steady-state position error because the vehicle loses control and reaches the workspace boundary in several evaluation episodes. Consistent with the training-return curves, the uncompressed variant achieves the best overall performance. Its larger parameter count provides greater representation and fitting capacity, allowing it to attain the lowest force-smoothness cost while remaining comparable to the best-performing variants across the other evaluation metrics. In Stage II, the base policy is frozen, and an adaptation module is trained through supervised distillation to imitate the teacher policy. As shown in Table~\ref{tab:ablation_metrics}, the student policy exhibits slightly inferior overall performance across the evaluation metrics compared with the teacher policy, with the uncompressed variant suffering the most pronounced performance degradation. In particular, its steady-state accuracy and settling-time performance deteriorate more noticeably. Therefore, although increasing the latent-space dimensionality yields the best performance in Stage I, the resulting high-dimensional representation is more difficult for the adaptation module to learn in Stage II, ultimately degrading the performance of the student policy. For the single-encoder variant, the absence of a structured, branch-specific latent-space architecture causes the time-invariant static parameters and time-varying dynamic states to become entangled in a shared latent representation, resulting in less stable latent encoding and a larger $Z$ MSE, which in turn leads to less smooth force outputs. Removing either the current-flow branch or the dynamic-state branch also degrades control performance. The current-flow branch primarily affects the settling time, whereas the dynamic-state branch has a greater influence on force smoothness. Removing the static-information branch leads to a slight degradation across all evaluation metrics. Overall, the proposed multi-encoder Privileged PPO architecture achieves the best balance across the evaluation metrics and delivers the strongest overall control performance. The ablation studies demonstrate the necessity and effectiveness of each architectural component.

The entire training process was conducted on a workstation equipped with a 14-core Intel(R) Xeon(R) Gold 6430 CPU and an NVIDIA RTX 4090 GPU. The training time is approximately 3 hours for Stage-I and 1 hours for Stage-II. During deployment, TSRCA-PPO requires 4.407 ms for one inference step, including 2.052 ms for the adaptation module and 2.015 ms for the policy actor. This latency is much shorter than the 16 ms control period, demonstrating that the proposed method satisfies the real-time requirement for online deployment.

\begin{table}[t]
	\centering
	\caption{Key implementation details and hyperparameters used in training phase I.}
	\label{tab:impl_hparams}
	\small
	\setlength{\tabcolsep}{5pt}
	\renewcommand{\arraystretch}{1.1}
	\begin{tabular}{ll}
		\hline
		Item & Value \\
		\hline
		GPU & NVIDIA GeForce RTX 4090 (1$\times$) \\
		
		Number of parallel environments & 4096 \\
		Episode length & 600 steps \\
		Simulation time step $\Delta t$ & 0.016 s \\
		Algorithm & PPO \\
		Max updates & 7628 \\
		Rollout horizon (\texttt{train\_every}) & 32 steps \\
		PPO epochs & 4 \\
		Minibatches per update & 16 \\
		Learning rate & $5\times10^{-4}$ (annealed) \\
		Target KL & 0.015 \\
		
		\hline
	\end{tabular}
\end{table}

\begin{table}[H]
	\centering
	\caption{Key implementation details and hyperparameters used in training phase II.}
	\label{tab:phase2_hparams}
	\setlength{\tabcolsep}{4pt}
	\renewcommand{\arraystretch}{1.1}
	\begin{tabular}{ll}
			\hline
			Item & Value \\
			\hline
			History length $L$ & 30 steps \\
			Number of parallel environments & 1024 \\
			Total frames & $1.0\times10^{8}$ \\
			Rollout horizon (\texttt{train\_every}) & 32 steps \\
			Max updates & 3050 \\
			Replay buffer size & 1000000 samples \\
			Learning rate & $10^{-3}$ with cosine annealing \\
			Mini-batch size $N_B$ & 80000 \\
			Epochs per replay update & 4 \\
			\hline
	\end{tabular}
\end{table}

\begin{table}[H]
	\centering
	\caption{Stage-I ablation and baseline evaluation results over 1,000 evaluation episodes in a 32-s station-keeping test.}
	\label{tab:stage1_ablation_metrics}
	\setlength{\tabcolsep}{2.5pt}
	\renewcommand{\arraystretch}{1.15}
	\resizebox{\linewidth}{!}{%
	\begin{tabular}{lccccccc}
		\toprule
		Method
		& $\mathrm{SS}_{\mathrm{pos}}$ (m)
		& $\mathrm{SS}_{\mathrm{att}}$ (deg)
		& $t_s$ (s)
		& Success (\%)
		& $J_E$ ($10^6$)
		& $J_{\mathrm{FS}}$ (N)
		& Inference (ms) \\
		\midrule
		\textbf{Multi-encoder Privileged PPO (proposed)} & $0.0063\pm0.0023$ & $0.5135\pm0.1667$ & 2.84 & 99.9 & 2.46 & 0.26 & 1.66 \\
		w/o static parameters & $0.0065\pm0.0024$ & $0.5236\pm0.1684$ & 2.77 & 99.9 & 2.49 & 0.26 & 1.47 \\
		w/o dynamic states & $0.0065\pm0.0024$ & $0.5970\pm0.2374$ & 2.80 & 99.8 & 2.46 & 0.61 & 1.48 \\
		w/o current parameters & $0.0062\pm0.0025$ & $0.4941\pm0.1601$ & 2.94 & 99.9 & 2.40 & 0.33 & 1.50 \\
		Uncompressed privileged information & $0.0065\pm0.0029$ & $0.5614\pm0.1876$ & 2.86 & 99.8 & 2.42 & 0.12 & 1.60 \\
		Single privileged encoder & $0.0066\pm0.0025$ & $0.5051\pm0.1591$ & 2.76 & 100.0 & 2.44 & 0.50 & 1.22 \\
		PPO baseline & $0.0093\pm0.0037$ & $0.8282\pm0.3107$ & 9.15 & 87.6 & 2.43 & 0.82 & 0.86 \\
		TQC baseline & $0.0150\pm0.2554$ & $1.8948\pm1.2507$ & 26.14 & 5.5 & 2.45 & 0.88 & 0.36 \\
		\bottomrule
	\end{tabular}%
	}
\end{table}

\FloatBarrier

\begin{table}[t]
	\centering
	\caption{Stage-II ablation and baseline evaluation results over 1,000 evaluation episodes in a 32-s station-keeping test.}
	\label{tab:ablation_metrics}
	\setlength{\tabcolsep}{2.5pt}
	\renewcommand{\arraystretch}{1.15}
	\resizebox{\linewidth}{!}{%
	\begin{tabular}{lcccccccc}
		\toprule
		Method
		& $\mathrm{SS}_{\mathrm{pos}}$ (m)
		& $\mathrm{SS}_{\mathrm{att}}$ (deg)
		& $t_s$ (s)
		& Success (\%)
		& $J_E$ ($10^6$)
		& $J_{\mathrm{FS}}$ (N)
		& $Z$ MSE
		& Inference (ms) \\
		\midrule
		\textbf{Multi-encoder Privileged PPO (proposed)} & $0.0070\pm0.0025$ & $0.5537\pm0.1637$ & 3.04 & 100.0 & 2.58 & 0.27 & 0.0340 & 4.41 \\
		w/o static parameters & $0.0070\pm0.0026$ & $0.5724\pm0.1702$ & 3.12 & 100.0 & 2.59 & 0.27 & 0.0302 & 4.16 \\
		w/o dynamic states & $0.0065\pm0.0024$ & $0.5947\pm0.2209$ & 3.00 & 100.0 & 2.57 & 0.62 & 0.0186 & 4.69 \\
		w/o current parameters & $0.0069\pm0.0026$ & $0.5206\pm0.1586$ & 3.18 & 99.8 & 2.51 & 0.35 & 0.0366 & 4.42 \\
		Uncompressed privileged information & $0.0080\pm0.0032$ & $0.6384\pm0.2049$ & 4.61 & 99.6 & 2.55 & 0.14 & 0.0477 & 5.38 \\
		Single privileged encoder & $0.0064\pm0.0022$ & $0.5266\pm0.1510$ & 2.82 & 100.0 & 2.55 & 0.52 & 0.1098 & 4.53 \\
		PPO baseline & $0.0093\pm0.0037$ & $0.8282\pm0.3107$ & 9.15 & 87.6 & 2.43 & 0.82 & -- & 0.86 \\
		TQC baseline & $0.0150\pm0.2554$ & $1.8948\pm1.2507$ & 26.14 & 5.5 & 2.45 & 0.88 & -- & 0.36 \\
		\bottomrule
	\end{tabular}%
	}
\end{table}
\FloatBarrier
\begin{figure}[H]
	\centering
	\includegraphics[width=\linewidth]{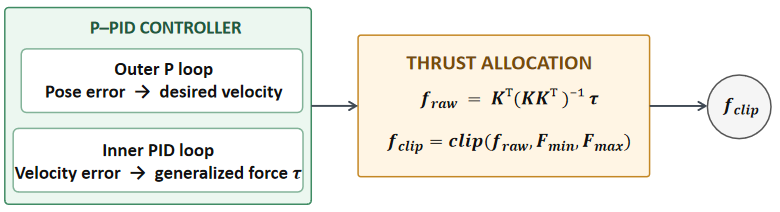}
	\caption{Architecture of the conventional cascaded P-PID controller using pseudoinverse thrust allocation.}
	\label{fig:ppid_architecture}
\end{figure}

\subsection{Comparison with Conventional Methods}
Figure~\ref{fig:ppid_architecture} illustrates the conventional cascaded P-PID control architecture used for comparison. In this section, we compare the proposed method TSRCA-PPO with a conventional cascaded P-PID control architecture, in which an outer-loop proportional controller generates the desired linear and angular velocities from the position and attitude errors, while an inner-loop PID controller computes the desired generalized forces and moments. These commands are subsequently distributed among the individual thrusters using a pseudoinverse thrust-allocation method, followed by element-wise clipping to the physical thrust limits. Table~\ref{tab:conventional_comparison} quantitatively compares the two methods across multiple performance metrics. The proposed TSRCA-PPO policy outperforms the P-PID controller across all reported metrics under the same settling criterion. Relative to P-PID, TSRCA-PPO reduces the position error by approximately 57.3\%, the attitude error by 23.5\%, and the energy-consumption proxy by 6.4\%. More importantly, it reduces the average per-thruster second difference by approximately 84.1\%, indicating substantially smoother thrust commands. The proposed method also shortens the mean settling time from 28.55~s to 3.04~s and increases the success rate from 72.8\% to 100.0\%, demonstrating a better overall balance among station-keeping accuracy, transient performance, energy efficiency, and thrust smoothness.

\begin{table}[H]
	\centering
	\caption{Comparison between the proposed TSRCA-PPO policy and the cascaded P-PID controller over 1,000 evaluation episodes in a 32-s station-keeping test. The settling criterion is $0.02$~m for position error and $2^\circ$ for attitude error.}
	\label{tab:conventional_comparison}
	\setlength{\tabcolsep}{3pt}
	\renewcommand{\arraystretch}{1.15}
	\resizebox{\linewidth}{!}{%
	\begin{tabular}{lcccccc}
		\toprule
		Method
		& Position error (m)
		& Attitude error (deg)
		& $t_s$ (s)
		& Success (\%)
		& $J_E$ ($10^6$)
		& $J_{\mathrm{FS}}$ (N) \\
		\midrule
		TSRCA-PPO  & $0.0070\pm0.0025$ & $0.5537\pm0.1637$ & 3.04 & 100.0 & 2.58 & 0.27 \\ P-PID & $0.0164\pm0.0097$ & $0.7242\pm0.3445$ & 28.55 & 72.8 & 2.76 & 1.70 \\
		\bottomrule
	\end{tabular}%
	}
\end{table}
\FloatBarrier

\begin{figure}[p]
	\centering
	\captionsetup[subfloat]{font=footnotesize}
	\subfloat[Position-error comparison.\label{fig:step_current_position_error}]{%
		\includegraphics[width=0.45\linewidth]{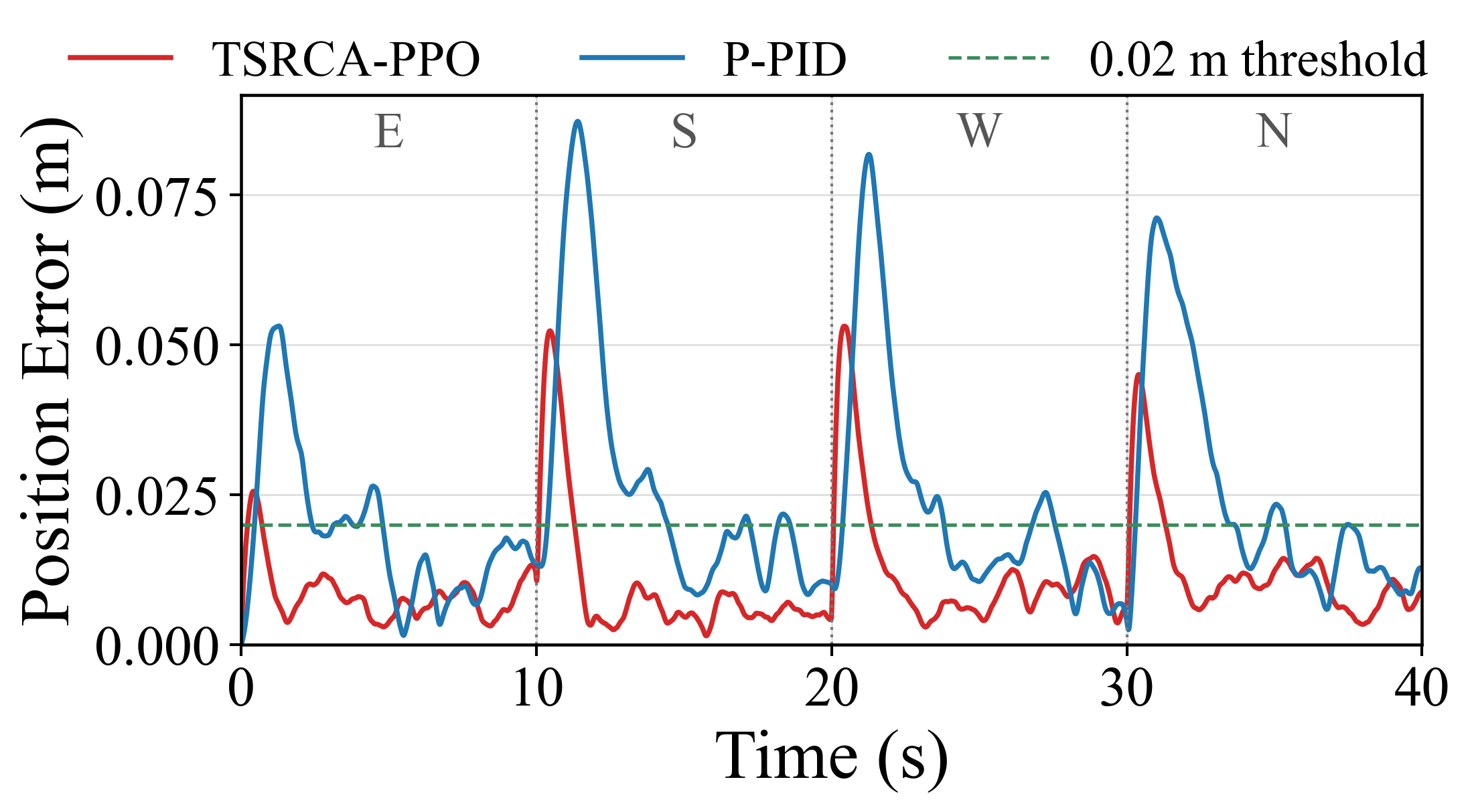}}
	\hfill
	\subfloat[Attitude-error comparison.\label{fig:step_current_attitude_error}]{%
		\includegraphics[width=0.45\linewidth]{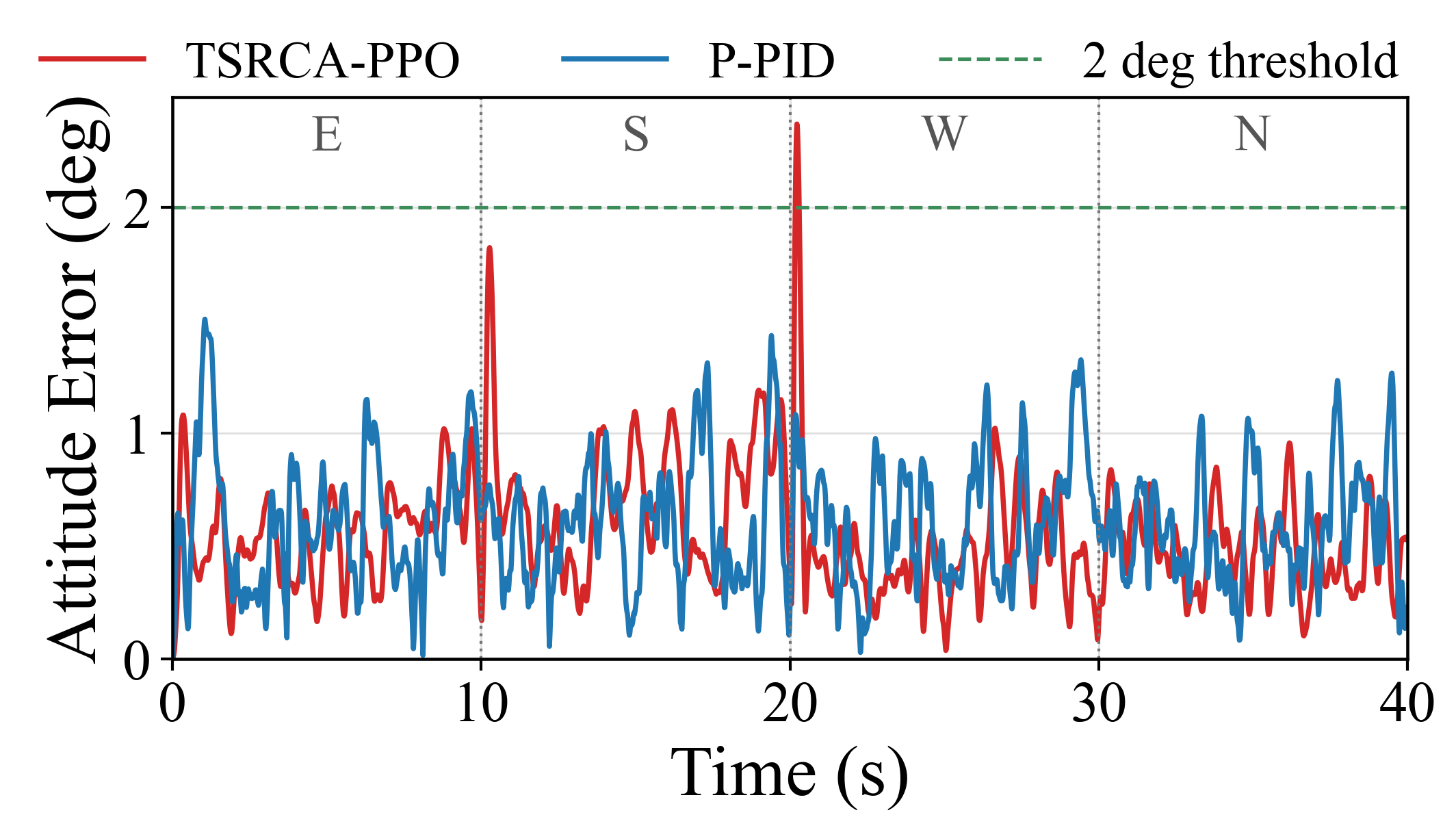}}

	\subfloat[Stepwise ocean-current input.\label{fig:step_current_input}]{%
		\includegraphics[width=0.45\linewidth]{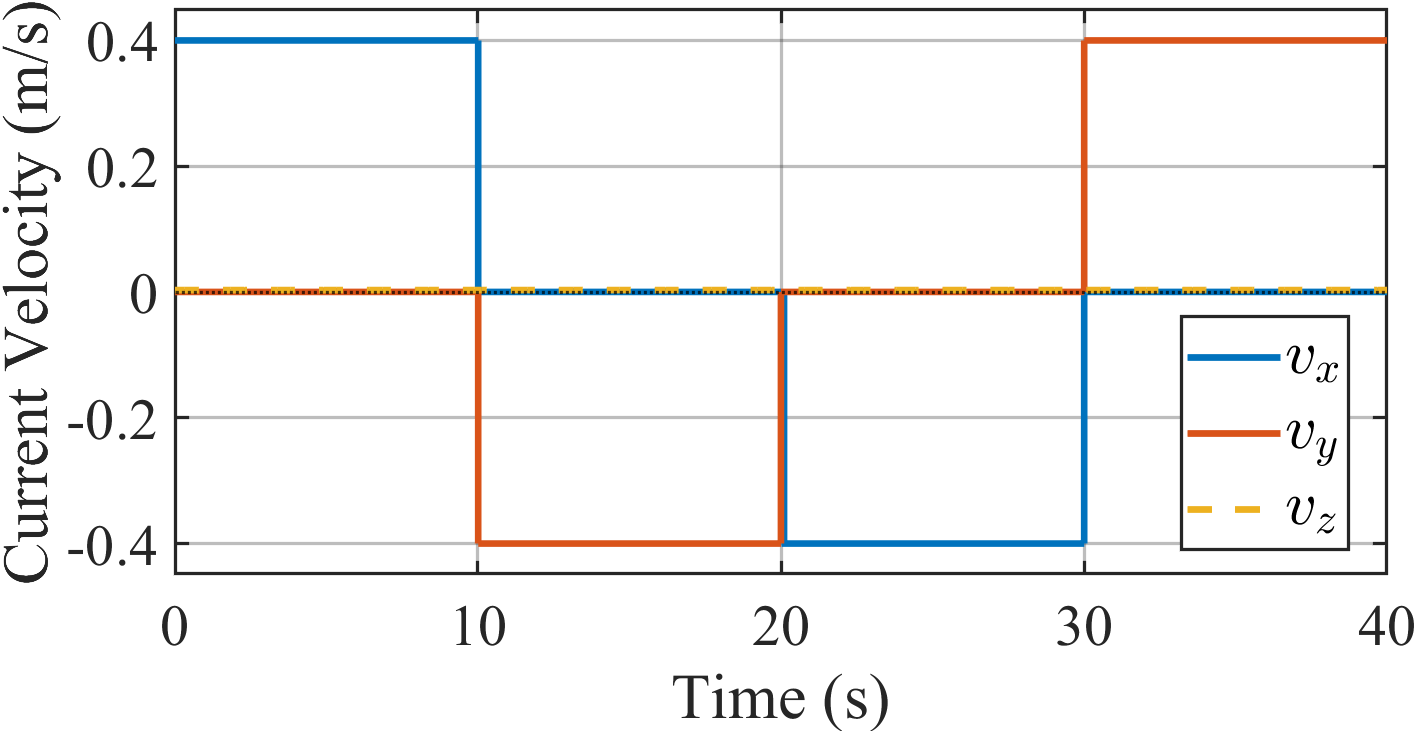}}
	\hfill
	\subfloat[Latent-feature estimation error.\label{fig:step_current_z_mse}]{%
		\includegraphics[width=0.45\linewidth]{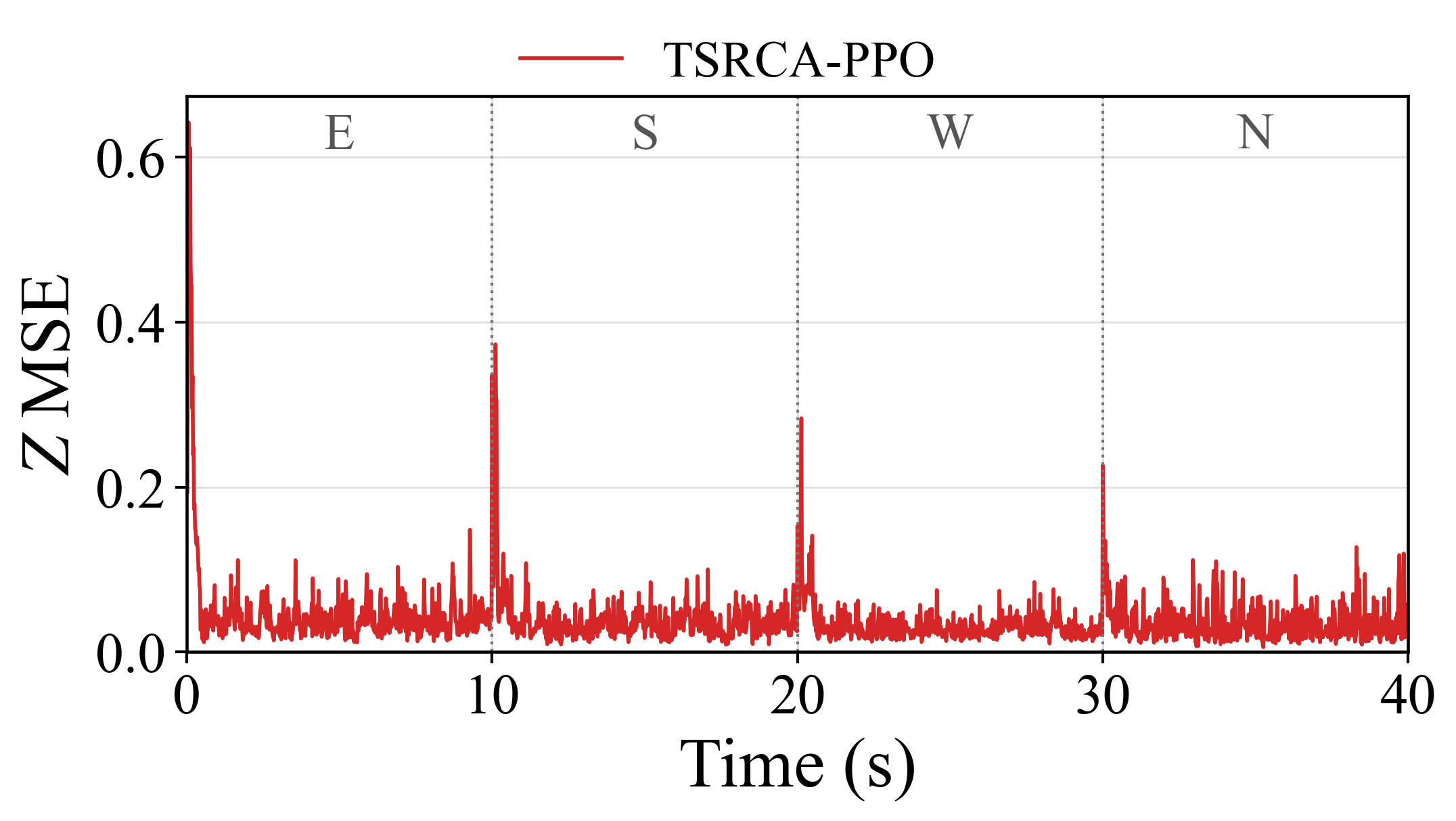}}
	\caption{Comparison between TSRCA-PPO and P-PID under stepwise east--south--west--north ocean-current disturbances: position and attitude errors, current input, and latent-feature estimation error.}
	\label{fig:step_current_comparison}
\end{figure}

\begin{figure}[p]
	\centering
	\includegraphics[width=0.90\linewidth]{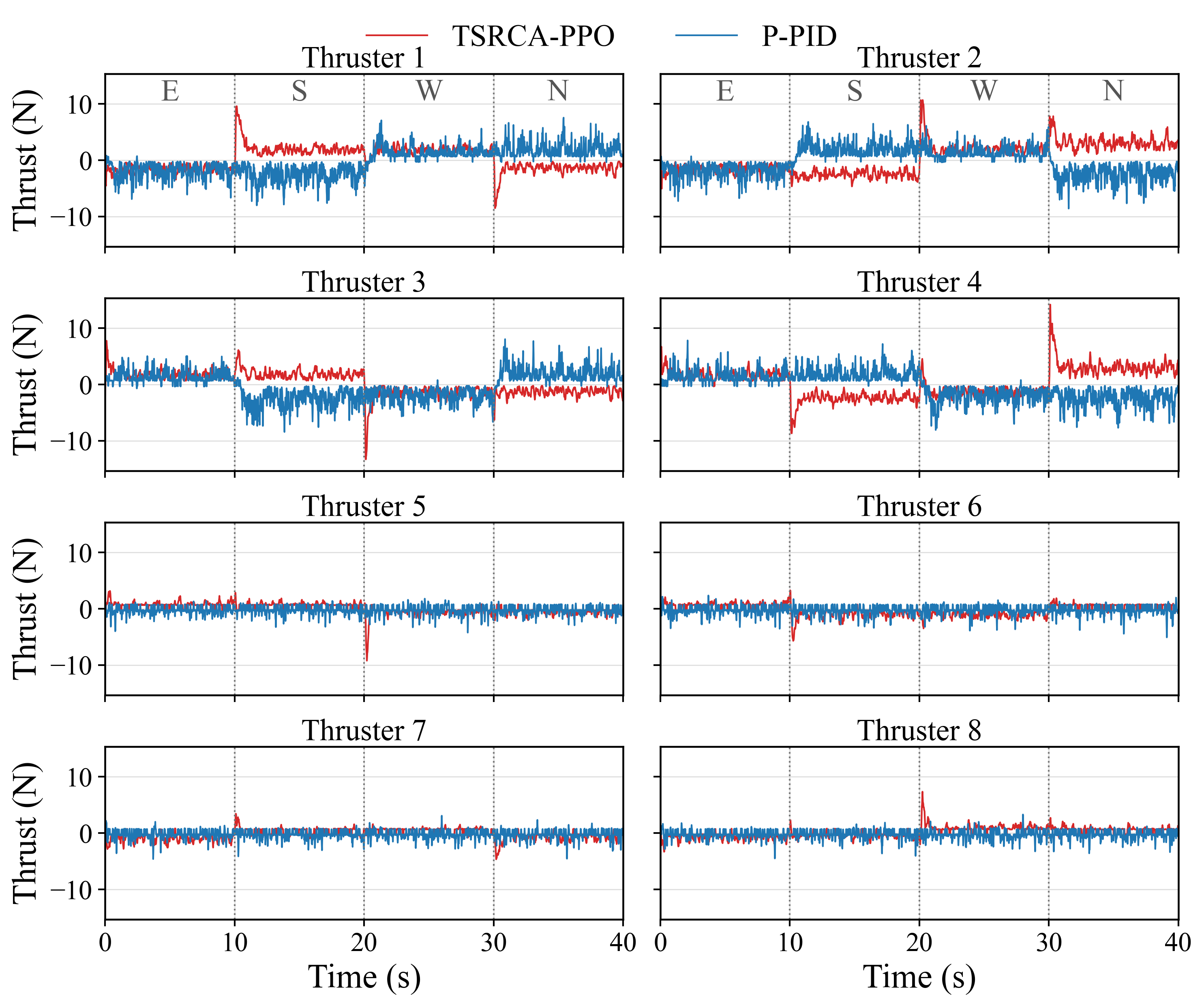}
	\caption{Individual-thruster force comparison between TSRCA-PPO and P-PID under stepwise east--south--west--north ocean-current disturbances.}
	\label{fig:step_current_thrust}
\end{figure}

Figures~\ref{fig:step_current_comparison} and~\ref{fig:step_current_thrust}
present a representative example in which the two algorithms perform a
station-keeping task under both the nominal model and a
\(0.4\,\mathrm{m/s}\) ocean current, with the vehicle commanded to maintain
a fixed position and attitude at the origin. The current direction is changed
stepwise to east, south, west, and north at \(t=0\), \(10\), \(20\), and
\(30\,\mathrm{s}\), respectively.
The experimental results are consistent with those reported in Table~\ref{tab:conventional_comparison}: TSRCA-PPO converges substantially faster than P-PID and achieves higher steady-state accuracy. The adaptation module of TSRCA-PPO rapidly reconstructs the current latent feature, providing greater sensitivity to changes in the ocean-current conditions and enabling the vehicle attitude to recover within a very short time after each current-direction transition. As shown in Fig.~\ref{fig:step_current_thrust}, the single-trajectory force-smoothness costs are $J_{\mathrm{FS}}=0.23$ for TSRCA-PPO and $J_{\mathrm{FS}}=0.92$ for P-PID. The substantially larger $J_{\mathrm{FS}}$ of P-PID reflects its more oscillatory thruster-force outputs, which impose greater mechanical stress on the actuators and may accelerate thruster wear.

\FloatBarrier
 
\section{Conclusions and Future work}
 This paper proposes an end-to-end reinforcement-learning-based position controller for remotely operated vehicles, named thrust smoothness rapid current adaptation proximal policy optimization (TSRCA-PPO), to achieve near-optimal control performance in the presence of ocean currents. The proposed method adopts a knowledge-distillation strategy. First, a teacher policy is trained using PPO with access to privileged information, enabling it to learn smooth, fast, energy-efficient and highly accurate thrust-allocation policies. Multiple encoders are introduced to extract structured feature representations from different types of privileged information. Then, on-policy trajectory data generated by the teacher policy are used to distill a student policy to infer the latent privileged information from historical onboard observations. After training, the student policy can be deployed to achieve near-optimal performance using only observation histories. Ablation studies validate the effectiveness and rationality of the proposed reward function and model architecture. Simulation results demonstrate that the proposed TSRCA-PPO method consistently
 outperforms the conventional cascaded P-PID controller across all evaluation
 metrics. Specifically, TSRCA-PPO reduces the position error, attitude error,
 settling time, energy index, and thrust-smoothness index
 to \(42.7\%\), \(76.5\%\), \(10.6\%\), \(93.5\%\), and \(15.9\%\) of the
 corresponding P-PID values, respectively.

Despite the substantial performance improvements achieved by the proposed method, physical experiments could not be conducted due to limitations in the available experimental facilities. Future work will focus on real-world experiments using physical underwater vehicles to further validate the method’s practical applicability and sim-to-real transfer performance.

\bibliographystyle{elsarticle-harv}
\bibliography{reference}
\end{document}